\documentclass{opt2025} 

\newtheorem{assumption}{Assumption}
\usepackage{wrapfig}
\title[Stepsize Shrinkage in Low-Precision Training]{Why Does Stochastic Gradient Descent Slow Down in Low-Precision Training?}

\optauthor{%
\Name{Vincent-Daniel Yun} \Email{juyoung.yun@usc.edu}\\
\addr University of Southern California, USA \\
\addr Open Neural Network Research Lab, MODULABS, Republic of Korea}

\begin{document}

\maketitle

\begin{abstract}%

Low-precision training has become crucial for reducing the computational and memory costs of large-scale deep learning. However, quantizing gradients introduces magnitude shrinkage, which can change how stochastic gradient descent (SGD) converges. In this study, we explore SGD convergence under a gradient shrinkage model, where each stochastic gradient is scaled by a factor \( q_k \in (0,1] \). We show that this shrinkage affect the usual stepsize \( \mu_k \) with an effective stepsize \( \mu_k q_k \), slowing convergence when \( q_{\min} < 1 \). With typical smoothness and bounded-variance assumptions, we prove that low-precision SGD still converges, but at a slower pace set by \( q_{\min} \), and with a higher steady error level due to quantization effects. We analyze theoretically how lower numerical precision slows training by treating it as gradient shrinkage within the standard SGD convergence setup.
\end{abstract}

\section{Introduction}
Deep learning models~\cite{deeplearning} have grown rapidly in size, while the amount of training data has increased exponentially with the development of the Internet~\cite{krizhevsky2012imagenet}. Training such large-scale models requires substantial GPU memory and computational resources~\cite{mixed}, motivating the widespread adoption of low-precision and mixed-precision training. Modern accelerators typically compute gradients in low precision (such as FP16, FP8, or FP4)~\cite{fp16,fp16-2,fp8-2,fp8-3,fp8-4} for speed and memory efficiency, while maintaining model parameters and accumulators in full precision (FP32). This hybrid approach is now standard practice in large-scale training pipelines but introduces an important numerical side effect. Low-precision gradients do not preserve their original magnitudes faithfully. Quantization maps many small values to coarse levels or zero, causing what we refer to as a \emph{systematic shrinkage} of gradient magnitudes during backpropagation. If \(g\) denotes the true full-precision gradient, the quantized gradient produced by FP16/FP8/FP4 arithmetic can be written as $\tilde{g} = q\,g + \varepsilon,$ where \(q \in (0,1]\) is a shrinkage factor and \(\varepsilon\) represents quantization noise. In practice, FP16 introduces only mild shrinkage, whereas lower precisions such as FP8 or FP4 exhibit substantially larger contraction of gradient magnitudes because of their limited dynamic range. Since optimizers apply the update \(w \leftarrow w - \mu \tilde{g}\), this shrinkage reduces the effective stepsize from \(\mu\) to $\mu_{\mathrm{eff}} = \mu q,$ thereby slowing optimization and increasing the steady-state error compared to FP32 training. Although convergence of low-precision SGD has been established in several works~\cite{lconv1,lconv2,lconv3}, the direct mathematical effect of shrinkage on the convergence rate has not been clearly isolated.

In this work, we explicitly incorporate the shrinkage factor \(q\) into the classical SGD convergence framework~\cite{sgd,sgd2,sgd3}. By tracking how \(q\) modifies the descent inequality, the second-moment bounds, and the resulting convergence rate, we show that low-precision arithmetic effectively rescales the stepsize and slows down training. Our analysis provides a simple theoretical explanation for the empirically observed slowdown in FP16/FP8/FP4 training and suggests how stepsize scheduling should be adapted in such low-precision regimes. A comprehensive overview of low-precision techniques is provided in Appendix~\ref{related}.

\paragraph{Our contribution.}
We analyze mixed-precision training under a gradient-shrinkage model and show that quantized gradients induce an effective stepsize $\mu q$ that slows convergence and enlarges the error floor. To highlight this central implication of our analysis, we formally state the following key observation:

\begin{center}
\textit{``Low-precision training reduces the effective stepsize to $\mu q$, thereby requiring a larger nominal stepsize to attain faster convergence.''}
\end{center}

By incorporating the shrinkage factor directly into the standard SGD convergence proof, we derive explicit bounds quantifying this slowdown and support the theory with short numerical experiments demonstrating the shrinkage in commonly used low-precision formats.

\section{Related Works}
\label{related}

Large neural network model has led to remarkable performance improvements~\cite{krizhevsky2012imagenet}, but also raised concerns over computational cost, energy efficiency, and accessibility. To address these, various compression and acceleration techniques have been proposed, including quantization~\cite{quant-nn,choi2018pact}, pruning~\cite{han2015learning,gale2019state}, and knowledge distillation~\cite{hinton2015distilling}. These approaches reduce model size, memory usage, and inference cost, often with minimal accuracy loss. 

Reducing the precision of weights, activations, and gradients is an effective way to cut memory and computation requirements~\cite{low-prec-sgd,dettmers2022llm}. Low-precision formats can be floating-point (e.g., FP16, FP8, FP4)~\cite{mixed,fp16,fp16-2,fp8-1,fp8-2,fp8-3,fp8-4,fp4} or fixed-point~\cite{fxpnet}. While fixed-point offers speed and memory benefits, it often suffers from limited dynamic range, especially for complex tasks~\cite{fp-cnn}. Mixed-precision training~\cite{mixed} combines low-precision computations with high-precision accumulations to maintain accuracy, and has been widely adopted in modern hardware, including NVIDIA GPUs and Google TPUs~\cite{koster2020bfloat16}. Specialized accelerators such as BitFusion~\cite{bitfusion} and FPGA-based solutions~\cite{fpga-dsp} further optimize low-precision execution.

\paragraph{Low-Precision Training Error.}
Prior work has attributed the accuracy degradation in low-precision training to several numerical effects. Earlier analyses~\cite{rel1} show that quantization introduces additional stochastic noise that limits statistical accuracy when left uncompensated. Subsequent studies~\cite{rel2} report that coarse quantization perturbs activation distributions, leading to bias shifts that are especially harmful in architectures with sensitive activation statistics. Other investigations~\cite{rel3} observe that low-bitwidth networks are more prone to suboptimal minima, suggesting that reduced precision inherently complicates optimization.

While these studies primarily focus on noise amplification, activation distortion, or architectural sensitivity, our work offers a complementary and distinct perspective. We show that low-precision arithmetic induces a \emph{systematic shrinkage} of gradient magnitudes, which directly reduces the effective stepsize during optimization. By incorporating this shrinkage into a standard SGD convergence framework, we provide a clean mathematical explanation for the slower convergence and elevated error floor observed in low-precision training.

\section{Problem Setup}
\paragraph{Notation.}
We follow the standard SGD convergence proof~\cite{sgd}. The expectation over all sources of randomness (data sampling and quantization) is written $\mathbb{E}[\cdot]$, while the conditional expectation given the $\sigma$-algebra $\mathcal{F}_k$ of all randomness up to iteration $k$ is $\mathbb{E}[\cdot \mid \mathcal{F}_k]$.  At iteration $k$, the stochastic gradient is $g(w_k, \xi_k) = \nabla F(w_k; \xi_k)$ and the low-precision gradient is $\tilde{g}(w_k, \xi_k) = q_k\,g(w_k, \xi_k) + \varepsilon_k,$ where the shrinkage factor $q_k \in [q_{\min}, q_{\max}] \subset (0,1]$ and the quantization noise $\varepsilon_k$ satisfies $\mathbb{E}[\varepsilon_k \mid \mathcal{F}_k] = 0$ and $\mathbb{E}[\| \varepsilon_k \|_2^2] \le \sigma_\varepsilon^2$.  If the nominal stepsize is $\mu_k > 0$, the effective stepsize is $\mu_k q_k$.

\begin{wrapfigure}{l}{0.5\textwidth}
    \centering
    \includegraphics[width=\linewidth]{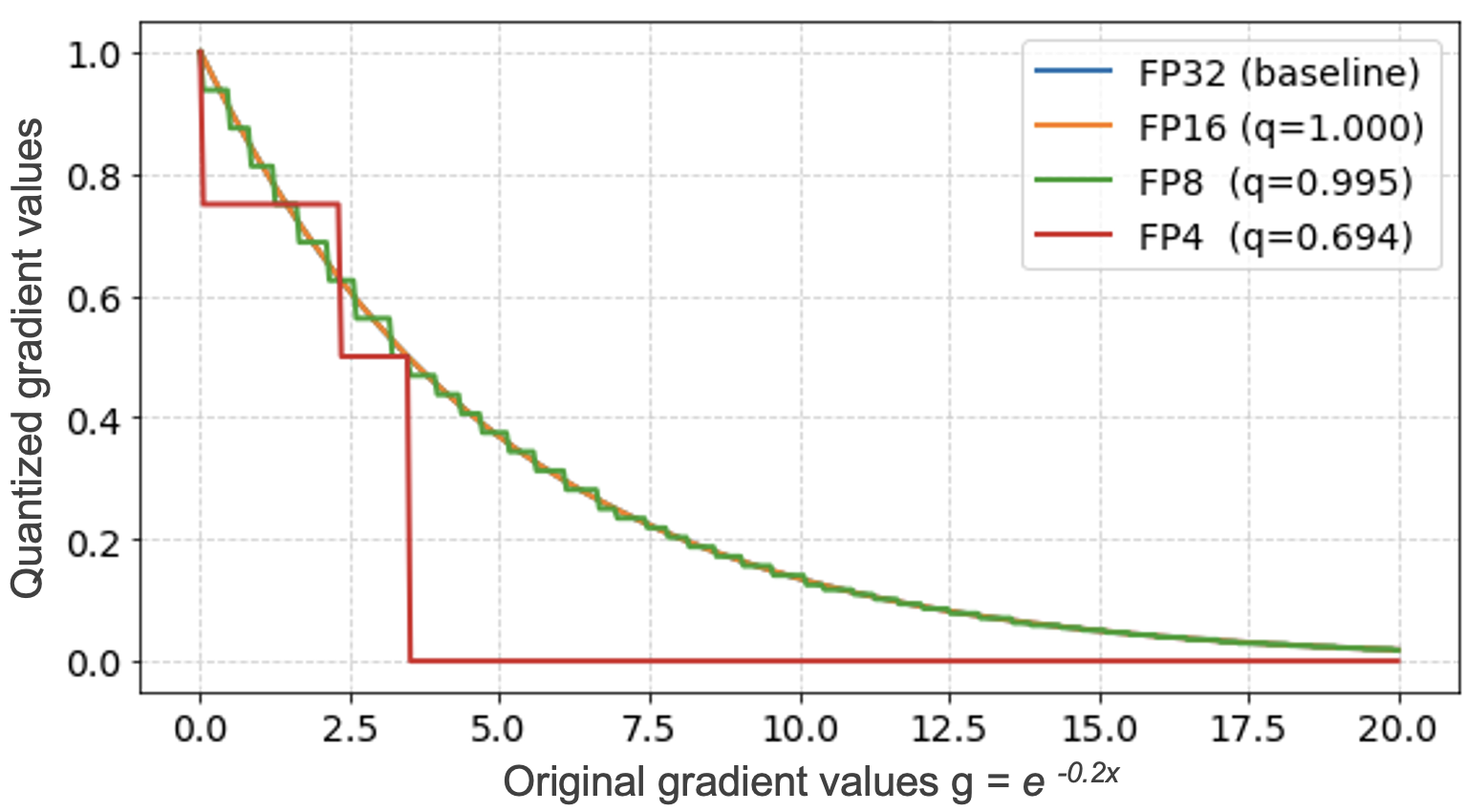}
    \vspace{-20pt}
    \caption{Quantization effect on a slowly decaying gradient-like function 
    $g = e^{-0.2x}$ without AMP or loss scaling.}
    \vspace{-15pt}
    \label{fig:problem}
\end{wrapfigure} 

\paragraph{Problem Setup.}
We minimize the expected loss $F(w) = \mathbb{E}_{\xi}[\ell(\xi, w)]$ over $w \in \mathbb{R}^d$, where $\xi$ is drawn from data distribution $\mathcal{D}$.  The optimization goal is to find $w_* \in \mathbb{R}^d$ such that $F(w_*) = \min_{w} F(w)$.  In the full-precision setting, SGD updates parameters via $w_{k+1} = w_k - \mu_k\, g(w_k, \xi_k),$ where $g(w_k, \xi_k) = \nabla F(w_k; \xi_k)$ and $\xi_k \stackrel{\text{i.i.d.}}{\sim} \mathcal{D}$. In low-precision formats (e.g., FP16, FP8, FP4), the update becomes $w_{k+1} = w_k - \mu_k q_k\, g(w_k, \xi_k) - \mu_k\, \varepsilon_k,$.  \\

Figure~\ref{fig:problem} illustrates this phenomenon for a smoothly decaying gradient-like signal $g$, showing how quantization maps many small values to zero or coarse levels, thereby reducing the overall magnitude.  From FP16 to FP4, the shrinkage factor $q$ decreases noticeably, and the gradient curve deviates more from the FP32 baseline.  The $q$ values were computed by measuring the ratio $\| \tilde{g} \|_2 / \| g \|_2$ after quantizing $g$ to each format without AMP or loss scaling.

Under standard convergence assumptions~\cite{sgd} but with the low-precision modifications above, the descent inequality effectively replaces $\mu_k$ by $\mu_k q_{\min}$, leading to slower convergence when $q_{\min} < 1$, while the noise term $\varepsilon_k$ adds extra variance to the error floor.  
In the next theoretical analysis section, we show that low-precision SGD still converges under these conditions by adapting a basic proof of SGD convergence~\cite{sgd}, and highlight how the stepsize shrinkage impacts the convergence speed.

\section{Theoretical Analysis}
We prove SGD convergence, showing that low-precision SGD converges more slowly, using two key ingredients: (i) smoothness of the objective and (ii) bounds on the first/second moments of the stochastic gradients $\{\tilde{g}(w_k,\xi_k)\}$ under the standard SGD proof of convergence~\cite{sgd}. Here, for notational simplicity, we denote $\mathbb{E}_{\xi_k, \mu_k, \varepsilon_k}[\cdot]$ by $\mathbb{E}_{\xi_k}[\cdot]$.

\begin{assumption}[Lipschitz-continuous objective gradients]
\label{assum1}
The objective $F : \mathbb{R}^d \to \mathbb{R}$ is continuously differentiable, and its gradient $\nabla F$ is $L$-Lipschitz continuous:
$\| \nabla F(w) - \nabla F(\bar{w}) \|_2 \le L \| w - \bar{w} \|_2$ where $\forall \{ w, \bar{w} \} \subset \mathbb{R}^d.$ This condition ensures that the gradient does not vary too rapidly with respect to $w$, a standard requirement for convergence analysis. A direct consequence is
\begin{equation}
    F(w) \le F(\bar{w}) + \nabla F(\bar{w})^\top (w - \bar{w}) + \frac{1}{2} L \| w - \bar{w} \|_2^2,
    \quad \forall \{ w, \bar{w} \} \subset \mathbb{R}^d. \label{eq1}
\end{equation}
\end{assumption}

\begin{assumption}[First and second moment limits with quantization]
\label{assum2}
The objective function and SGD satisfy the following: \\
\textbf{(a)} The sequence of iterates $\{ w_k \}$ is contained in an open set over which $F$ is bounded below by a scalar $F_{\inf}$. This requires $F$ to be bounded below in the region of iterates.\\
\textbf{(b)} There exist scalars $\mu_G \ge \mu > 0$ and $q_{\min} > 0$ such that, for all $k \in \mathbb{N}$. This ensures $-\tilde{g}(w_k, \xi_k)$ is a sufficient descent direction with magnitude comparable to $\nabla F(w_k)$ but reduced by $q_{\min}$,
    \begin{align}
        \nabla F(w_k)^\top \mathbb{E}_{\xi_k}[\tilde{g}(w_k, \xi_k)] &\ge q_{\min}\mu \| \nabla F(w_k) \|_2^2, \label{eq7} \\
        \| \mathbb{E}_{\xi_k}[\tilde{g}(w_k, \xi_k)] \|_2 &\le q_{\max}\mu_G \| \nabla F(w_k) \|_2. \label{eq8}
    \end{align}
\textbf{(c)} There exist scalars $M \ge 0$ and $M_V \ge 0$ such that, for all $k \in \mathbb{N}$, where $\tilde{M} := q_{\max}^2 M + M_\varepsilon$ and $\tilde{M}_V := q_{\max}^2 M_V + M_{\varepsilon,V}$ account for quantization noise. This bounds the variance of $\tilde{g}(w_k, \xi_k)$, allowing it to be nonzero at stationary points and grow quadratically for convex quadratics
\begin{equation}
    \mathbb{V}_{\xi_k,q_k,\varepsilon_k} \| \tilde{g}(w_k, \xi_k) \|_2 \le \tilde{M} + \tilde{M}_V \| \nabla F(w_k) \|_2^2, \label{eq9}
\end{equation}
\end{assumption}

\begin{assumption}[Strong convexity]
\label{assum3}
The objective $F : \mathbb{R}^d \to \mathbb{R}$ is strongly convex: there exists $c > 0$ such that
\begin{equation}
    F(\overline{w}) \ge F(w) + \nabla F(w)^\top (\overline{w} - w) + \frac{1}{2} c \| \overline{w} - w \|_2^2, \label{eq15}
\end{equation}
for all $(\overline{w}, w) \in \mathbb{R}^d \times \mathbb{R}^d$.  
This implies $F$ has a unique minimizer $w_* \in \mathbb{R}^d$ with $F_* := F(w_*)$, and
\begin{equation}
    2c \left( F(w) - F_* \right) \le \| \nabla F(w) \|_2^2
    \quad \text{for all } w \in \mathbb{R}^d. \label{eq16}
\end{equation}
\end{assumption}

\begin{lemma}
\label{Lemma1}
Under Assumption~\ref{assum1}, the iterates of SGD with low-precision gradient $\tilde{g}(w_k,\xi_k)$ satisfy, for all $k \in \mathbb{N}$, 
\begin{equation}
    \mathbb{E}_{\xi_k}[F(w_{k+1})] - F(w_k)
    \le -\alpha_k \nabla F(w_k)^\top \mathbb{E}_{\xi_k}[\tilde{g}(w_k, \xi_k)] 
    + \frac{1}{2} \alpha_k^2 L \, \mathbb{E}_{\xi_k}[\| \tilde{g}(w_k, \xi_k) \|_2^2]. \label{eq2}
\end{equation}
\end{lemma}

\begin{lemma}
\label{lemma2}
Under Assumptions~\ref{assum1} and~\ref{assum2}, the iterates of SGD satisfy, for all $k \in \mathbb{N}$,
\begin{align}
    \mathbb{E}_{\xi_k}[F(w_{k+1})] - F(w_k) 
    &\le \underbrace{-q_{\min}\mu \alpha_k \| \nabla F(w_k)}_{\text{stepsize shrinkage}} \|_2^2 
        + \frac{1}{2} \alpha_k^2 L \, \mathbb{E}_{\xi_k}[\| \tilde{g}(w_k, \xi_k) \|_2^2],  \label{eq11} \\
    &\le -\underbrace{(q_{\min}\mu - \frac{1}{2} \alpha_k L \tilde{M}_G )}_{\text{stepsize shrinkage}} \alpha_k \| \nabla F(w_k) \|_2^2 
        + \frac{1}{2} \alpha_k^2 L \tilde{M}. \label{eq12}
\end{align}
\end{lemma}

\begin{theorem}[Strongly Convex Objective, Fixed Stepsize with Quantization]
\label{theory1}
Under Assumptions~\ref{assum1}, \ref{assum2}, and~\ref{assum3} with $F_{\inf} = F_*$, we suppose SGD uses the low-precision gradient $\tilde{g}(w_k,\xi_k)$  satisfying $0 < \bar{\alpha} \le \frac{\mu_q}{L \tilde{M}_G}, $ where $\mu_q := q_{\min} \mu.$
Then, for all $k \in \mathbb{N}$,
\begin{align}
    \mathbb{E}[F(w_k) - F_*] &\le \frac{\bar{\alpha} L \tilde{M}}{2 c \mu_q}
    + \underbrace{( 1 - \bar{\alpha} c \mu_q )^{k-1}}_{\text{stepsize shrinkage}} \left( F(w_1) - F_* - \frac{\bar{\alpha} L \tilde{M}}{2 c \mu_q} \right),\label{eq17}
    \\
    &\quad \underset{k \to \infty}
    {\longrightarrow} \quad
    \frac{\bar{\alpha} L \tilde{M}}{2 c \mu_q} = \frac{1}{q_{min}}\frac{\bar{\alpha} L \tilde{M}}{2 c\mu} 
\end{align}
\end{theorem}
Note that when $q_{\min} < 1$, the reduced factor $\mu_q = q_{\min} \mu$ makes $(1 - \bar{\alpha} c \mu_q)^{k-1}$ decay at a slower rate, thereby reducing the convergence speed. Specifically, since $\mu_q$ appears in the denominator of the limit term $\frac{\bar{\alpha} L \tilde{M}}{2 c \mu_q}$, a smaller $\mu_q$ increases this term, leading to a larger asymptotic error bound.

\noindent For stepsizes $\alpha_r = \alpha_1 2^{-r}$, the bound is
$\mathbb{E}[F(w_{k_r+1}) - F_*] \le 2F_{\alpha_r}$ and $ F_{\alpha_r} := \frac{\alpha_r L \tilde{M}}{2 c \mu_q}$ requiring $k_{r+1} - k_r \approx \frac{\log 3}{\alpha_r c \mu_q} = \mathcal{O}(2^r),$
with the Robbins–Monro condition $\sum_{k=1}^\infty \alpha_k = \infty$ and $\sum_{k=1}^\infty \alpha_k^2 < \infty.$

\begin{theorem}[Strongly Convex Objective, Diminishing Stepsizes with Quantization]
\label{theorem2}
Under Assumptions~\ref{assum1}, \ref{assum2}, and~\ref{assum3} (with $F_{\mathrm{inf}} = F_*$), suppose that SGD with the low-precision gradient $\tilde{g}(w_k,\xi_k)$ is run with a stepsize sequence $\alpha_k = \frac{\beta}{\gamma + k}, \beta > \frac{1}{c \mu_q}, \gamma > 0,$ such that $\alpha_1 \le \frac{\mu_q}{L \tilde{M}_G}$, where $\mu_q := q_{\min} \mu$.  
Then, for all $k \in \mathbb{N}$, the expected optimality gap satisfies
\begin{equation}
    \mathbb{E}[F(w_k) - F_*] \le \frac{\nu_q}{\gamma + k},
    \label{eq25}
\end{equation}
where $\nu_q := \max \left\{ \frac{\beta^2 L \tilde{M}}{2 (\beta c \mu_q - 1)}, \; (\gamma + 1)(F(w_1) - F_*) \right\}.$
Since $\mu_q$ appears in the denominator of $\frac{\beta^2 L \tilde{M}}{2 (\beta c \mu_q - 1)}$, a smaller $\mu_q$ reduces the denominator and thus increases the bound.
\end{theorem}

Based on Theorem~\ref{theorem2}, when $q_{\min} < 1$, the reduced effective coefficient $\mu_q = q_{\min} \mu$ makes the $O(1/k)$ convergence rate slower and increases the constant factor in the bound.

\begin{remark}[Convergence Under Quantization]
Even though quantization introduces the shrinkage factor $q_{\min} < 1$ and reduces the effective coefficient $\mu_q = q_{\min}\mu$, which slows the $O(1/k)$ convergence rate, Theorem~\ref{theorem2} still guarantees convergence. Since the bound $\nu_q / (\gamma + k)$ tends to zero as $k \to \infty$, the optimality gap satisfies $\mathbb{E}[F(w_k) - F_*] \to 0.$ Thus, low-precision SGD continues to converge to the optimum, with the shrinkage factor influencing only the speed of convergence.
\end{remark}

\paragraph{Comparison with full precision.}
In the full-precision case ($q_k\equiv 1$, $\varepsilon_k\equiv 0$), the same argument yields $\mathbb{E}[F(w_k)-F_*] \le \frac{\nu}{\gamma+k} $ and $
\nu := \max\!\left\{ \frac{\beta^2 L M}{2(\beta c \mu - 1)},\; (\gamma+1)(F(w_1)-F_*) \right\}.    $
Relating the noise-dominated terms, we define the inflation factor
$\rho \;:=\; \frac{\tilde M}{M}\cdot \frac{\beta c \mu - 1}{\,\beta c \mu_q - 1\,}(>0),$ so that is $\frac{\beta^2 L \tilde M}{2(\beta c \mu_q - 1)} \;=\; \rho \cdot \frac{\beta^2 L M}{2(\beta c \mu - 1)}.$
Consequently where $A:=\frac{\beta^2 L M}{2(\beta c \mu - 1)},\;\; B:=(\gamma+1)(F(w_1)-F_*)$,
\begin{equation}
\nu_q \;=\; \max\{\rho A,\;B\}, \nu \;=\; \max\{A,\;B\},
\end{equation}
In particular, if $\rho\ge 1$ (e.g., when $q_{\min}<1$ and $\tilde M$ is sufficiently large relative to $M$ so that $\tilde M/M \ge \tfrac{\beta c \mu_q - 1}{\beta c \mu - 1}$), then $\nu_q \ge \nu$ and thus $\frac{\nu}{\gamma+k} \;\le\; \frac{\nu_q}{\gamma+k}\qquad\text{for all }k\in\mathbb{N}.$
This makes explicit that quantization (via $\mu_q$ and $\tilde M$) weakens the bound compared to full precision.


\section{Conclusion}
Our analysis establishes that the slowdown in low-precision SGD arises fundamentally from the gradient shrinkage factor $q_{\min} < 1$. By incorporating this factor into the descent inequality and the moment bounds, we show that the effective stepsize becomes $\alpha_k \mu_q$ with $\mu_q = q_{\min}\mu$, directly scaling down the rate at which the iterates approach the optimum. This modification yields slower convergence in both the fixed-stepsize and diminishing-stepsize regimes. Our proof follows the classical SGD convergence framework and isolates a simple and explicit mechanism, namely stepsize shrinkage, that explains why lower numerical precision leads to reduced optimization speed. Together, these theoretical findings clarify the role of gradient shrinkage in low-precision optimization and offer practical guidance for selecting stepsizes in low precision training.

\section{Acknowledgement}
This research was supported by Brian Impact Foundation, a non-profit organization dedicated to the advancement of science and technology for all.

\bibliography{sample}
\newpage
\appendix

\renewcommand{\thetheorem}{\arabic{theorem}}
\renewcommand{\thelemma}{\arabic{lemma}}
\renewcommand{\theassumption}{\arabic{assumption}}
\setcounter{theorem}{0}
\setcounter{assumption}{0}

\begin{center}
{\LARGE \textbf{Appendix}}\\[1em]
\end{center}

\begin{assumption}[Lipschitz-continuous objective gradients]
\label{assum1}
The objective $F : \mathbb{R}^d \to \mathbb{R}$ is continuously differentiable, and its gradient $\nabla F$ is $L$-Lipschitz continuous:
$\| \nabla F(w) - \nabla F(\bar{w}) \|_2 \le L \| w - \bar{w} \|_2$ where $\forall \{ w, \bar{w} \} \subset \mathbb{R}^d.$ This condition ensures that the gradient does not vary too rapidly with respect to $w$, a standard requirement for convergence analysis. A direct consequence is
\begin{equation}
    F(w) \le F(\bar{w}) + \nabla F(\bar{w})^\top (w - \bar{w}) + \frac{1}{2} L \| w - \bar{w} \|_2^2,
    \quad \forall \{ w, \bar{w} \} \subset \mathbb{R}^d. \label{eq1}
\end{equation}
\end{assumption}

\begin{assumption}[First and second moment limits with quantization]
\label{assum2}
The objective function and SGD satisfy the following: \\
\textbf{(a)} The sequence of iterates $\{ w_k \}$ is contained in an open set over which $F$ is bounded below by a scalar $F_{\inf}$. This requires $F$ to be bounded below in the region of iterates.\\
\textbf{(b)} There exist scalars $\mu_G \ge \mu > 0$ and $q_{\min} > 0$ such that, for all $k \in \mathbb{N}$. This ensures $-\tilde{g}(w_k, \xi_k)$ is a sufficient descent direction with magnitude comparable to $\nabla F(w_k)$ but reduced by $q_{\min}$,
    \begin{align}
        \nabla F(w_k)^\top \mathbb{E}_{\xi_k}[\tilde{g}(w_k, \xi_k)] &\ge q_{\min}\mu \| \nabla F(w_k) \|_2^2, \label{eq7} \\
        \| \mathbb{E}_{\xi_k}[\tilde{g}(w_k, \xi_k)] \|_2 &\le q_{\max}\mu_G \| \nabla F(w_k) \|_2. \label{eq8}
    \end{align}
\textbf{(c)} There exist scalars $M \ge 0$ and $M_V \ge 0$ such that, for all $k \in \mathbb{N}$, where $\tilde{M} := q_{\max}^2 M + M_\varepsilon$ and $\tilde{M}_V := q_{\max}^2 M_V + M_{\varepsilon,V}$ account for quantization noise. This bounds the variance of $\tilde{g}(w_k, \xi_k)$, allowing it to be nonzero at stationary points and grow quadratically for convex quadratics
\begin{equation}
    \mathbb{V}_{\xi_k,q_k,\varepsilon_k} \| \tilde{g}(w_k, \xi_k) \|_2 \le \tilde{M} + \tilde{M}_V \| \nabla F(w_k) \|_2^2, \label{eq9}
\end{equation}
\end{assumption}

\begin{assumption}[Strong convexity]
\label{assum3}
The objective $F : \mathbb{R}^d \to \mathbb{R}$ is strongly convex: there exists $c > 0$ such that
\begin{equation}
    F(\overline{w}) \ge F(w) + \nabla F(w)^\top (\overline{w} - w) + \frac{1}{2} c \| \overline{w} - w \|_2^2, \label{eq15}
\end{equation}
for all $(\overline{w}, w) \in \mathbb{R}^d \times \mathbb{R}^d$.  
This implies $F$ has a unique minimizer $w_* \in \mathbb{R}^d$ with $F_* := F(w_*)$, and
\begin{equation}
    2c \left( F(w) - F_* \right) \le \| \nabla F(w) \|_2^2
    \quad \text{for all } w \in \mathbb{R}^d. \label{eq16}
\end{equation}
\end{assumption}

\begin{lemma}
\label{Lemma1}
Under Assumption~\ref{assum1}, the iterates of SGD with low-precision gradient $\tilde{g}(w_k,\xi_k)$ satisfy, for all $k \in \mathbb{N}$, 
\begin{equation}
    \mathbb{E}_{\xi_k}[F(w_{k+1})] - F(w_k)
    \le -\alpha_k \nabla F(w_k)^\top \mathbb{E}_{\xi_k}[\tilde{g}(w_k, \xi_k)] 
    + \frac{1}{2} \alpha_k^2 L \, \mathbb{E}_{\xi_k}[\| \tilde{g}(w_k, \xi_k) \|_2^2]. \label{eq2}
\end{equation}
\end{lemma}

\begin{proof}
From Assumption~\ref{assum1},
\begin{align}
    F(w_{k+1}) - F(w_k)
    &\le \nabla F(w_k)^\top (w_{k+1} - w_k) + \frac{1}{2} L \| w_{k+1} - w_k \|_2^2 \label{eq3}\\
    &\le -\alpha_k \nabla F(w_k)^\top \big(\underbrace{q_k\,g(w_k, \xi_k) + \varepsilon_k}_{\text{stepsize shrinkage}}\big) 
        + \frac{1}{2} \alpha_k^2 L \, \| \underbrace{q_k\,g(w_k, \xi_k) + \varepsilon_k}_{\text{stepsize shrinkage}} \|_2^2. \label{eq4}
\end{align}
Taking expectations over $(\xi_k, q_k, \varepsilon_k)$, with $w_k$ fixed, yields \eqref{eq2}.
\end{proof}

This bound expresses the expected one-step change as the sum of a descent term and a curvature-dependent penalty, both influenced by $q_k$ and $\varepsilon_k$. If $\tilde{g}(w_k,\xi_k)$ is unbiased, then
\begin{equation}
    \mathbb{E}_{\xi_k}[F(w_{k+1})] - F(w_k) 
    \le -\alpha_k \| \nabla F(w_k) \|_2^2 
        + \frac{1}{2} \alpha_k^2 L \, \mathbb{E}_{\xi_k}[\| \underbrace{q_k\,g(w_k, \xi_k) + \varepsilon_k}_{\text{stepsize shrinkage}} \|_2^2]. \label{eq5}
\end{equation}

We guarantee SGD convergence when the stochastic directions and stepsizes make the right-hand side of (\ref{eq2}) bounded by a deterministic term that ensures sufficient descent in $F$. This requires constraints on the first and second moments of $\{ \tilde{g}(w_k, \xi_k) \}$ to limit the effect of the last term in (\ref{eq5}). We restrict the variance of $\tilde{g}$ as $\mathbb{V}_{\xi_k,q_k,\varepsilon_k} [\tilde{g}(w_k, \xi_k)] := \mathbb{E} \left[ \| \tilde{g}(w_k, \xi_k) \|_2^2 \right] - \| \mathbb{E} [\tilde{g}(w_k, \xi_k)] \|_2^2.$ \\

\noindent Together with the variance of $\tilde{g}$, these give the second moment bound:
\begin{equation}
    \mathbb{E}_{\xi_k} \left[ \| \tilde{g}(w_k, \xi_k) \|_2^2 \right] \le \tilde{M} + \tilde{M}_G \| \nabla F(w_k) \|_2^2 \quad \tilde{M}_G := \tilde{M}_V + q_{\max}^2\mu_G^2 \ge (q_{\min}\mu)^2 > 0.\label{eq10}
\end{equation}

The next lemma extends Lemma~\ref{Lemma1} under Assumption~\ref{assum2}.
\begin{lemma}
\label{lemma2}
Under Assumptions~\ref{assum1} and~\ref{assum2}, the iterates of SGD satisfy, for all $k \in \mathbb{N}$,
\begin{align}
    \mathbb{E}_{\xi_k}[F(w_{k+1})] - F(w_k) 
    &\le \underbrace{-q_{\min}\mu \alpha_k \| \nabla F(w_k)}_{\text{stepsize shrinkage}} \|_2^2 
        + \frac{1}{2} \alpha_k^2 L \, \mathbb{E}_{\xi_k}[\| \tilde{g}(w_k, \xi_k) \|_2^2],  \label{eq11} \\
    &\le -\underbrace{(q_{\min}\mu - \frac{1}{2} \alpha_k L \tilde{M}_G )}_{\text{stepsize shrinkage}} \alpha_k \| \nabla F(w_k) \|_2^2 
        + \frac{1}{2} \alpha_k^2 L \tilde{M}. \label{eq12}
\end{align}
\end{lemma}

\begin{proof}
From Lemma~\ref{Lemma1} with $\tilde{g}$, we have
\begin{align}
    \mathbb{E}_{\xi_k} \left[ F(w_{k+1}) \right] - F(w_k)
    &\le -\alpha_k \nabla F(w_k)^\top \mathbb{E}_{\xi_k} \left[ \tilde{g}(w_k, \xi_k) \right]
        + \frac{1}{2} \alpha_k^2 L \, \mathbb{E}_{\xi_k} \left[ \| \tilde{g}(w_k, \xi_k) \|_2^2 \right] \label{eq13} \\
    &\le \underbrace{-q_{\min}\mu \alpha_k \| \nabla F(w_k) \|_2^2}_{\text{stepsize shrinkage}}
        + \frac{1}{2} \alpha_k^2 L \, \mathbb{E}_{\xi_k} \left[ \| \tilde{g}(w_k, \xi_k) \|_2^2 \right], \label{eq14}
\end{align}
which yields \eqref{eq11}.  
Applying Assumption~\ref{assum2}(c) and the bound in \eqref{eq10} gives \eqref{eq12}.
\end{proof}

\begin{theorem}[Strongly Convex Objective, Fixed Stepsize with Quantization]
\label{theory1}
Under Assumptions~\ref{assum1}, \ref{assum2}, and~\ref{assum3} with $F_{\inf} = F_*$, we suppose SGD uses the low-precision gradient $\tilde{g}(w_k,\xi_k)$  satisfying $0 < \bar{\alpha} \le \frac{\mu_q}{L \tilde{M}_G}, $ where $\mu_q := q_{\min} \mu.$
Then, for all $k \in \mathbb{N}$,
\begin{align}
    \mathbb{E}[F(w_k) - F_*] &\le \frac{\bar{\alpha} L \tilde{M}}{2 c \mu_q}
    + \underbrace{( 1 - \bar{\alpha} c \mu_q )^{k-1}}_{\text{stepsize shrinkage}} \left( F(w_1) - F_* - \frac{\bar{\alpha} L \tilde{M}}{2 c \mu_q} \right),\label{eq17}
    \\
    &\quad \underset{k \to \infty}
    {\longrightarrow} \quad
    \frac{\bar{\alpha} L \tilde{M}}{2 c \mu_q} = \frac{1}{q_{min}}\frac{\bar{\alpha} L \tilde{M}}{2 c\mu} 
\end{align}
\end{theorem}
Note that when $q_{\min} < 1$, the reduced factor $\mu_q = q_{\min} \mu$ makes $(1 - \bar{\alpha} c \mu_q)^{k-1}$ decay at a slower rate, thereby reducing the convergence speed. Specifically, since $\mu_q$ appears in the denominator of the limit term $\frac{\bar{\alpha} L \tilde{M}}{2 c \mu_q}$, a smaller $\mu_q$ increases this term, leading to a larger asymptotic error bound.
\begin{proof}
From Lemma~\ref{lemma2}, for all $k \in \mathbb{N}$ we have
\begin{align}
    \mathbb{E}_{\xi_k}[F(w_{k+1})] - F(w_k)
    &\le -\left( \mu_q - \frac{1}{2} \bar{\alpha} L \tilde{M}_G \right) \bar{\alpha} \| \nabla F(w_k) \|_2^2
        + \frac{1}{2} \bar{\alpha}^2 L \tilde{M} \label{eq18}\\
    &\le -\frac{1}{2} \bar{\alpha} \mu_q \| \nabla F(w_k) \|_2^2
        + \frac{1}{2} \bar{\alpha}^2 L \tilde{M} \label{eq19}\\
    &\le -\bar{\alpha} c \mu_q \left( F(w_k) - F_* \right)
        + \frac{1}{2} \bar{\alpha}^2 L \tilde{M}, \label{eq20}
\end{align}
where \eqref{eq20} follows from the strong convexity bound \eqref{eq16}.
Subtracting $F_*$ and taking expectations gives $\mathbb{E}[F(w_{k+1}) - F_*] \le (1 - \bar{\alpha} c \mu_q) \, \mathbb{E}[F(w_k) - F_*] + \frac{1}{2} \bar{\alpha}^2 L \tilde{M}. $
Subtracting $\frac{\bar{\alpha} L \tilde{M}}{2 c \mu_q}$ from both sides yields
\begin{align}
    \mathbb{E}[F(w_{k+1}) - F_*] - \frac{\bar{\alpha} L \tilde{M}}{2 c \mu_q} 
    &= (1 - \bar{\alpha} c \mu_q) \left( \mathbb{E}[F(w_k) - F_*] - \frac{\bar{\alpha} L \tilde{M}}{2 c \mu_q} \right). \label{eq23}
\end{align}
Since $0 < \bar{\alpha} c \mu_q \le \frac{c\mu_q^2}{LM_G} \le \frac{c\mu_q^2}{L\mu_k} = \frac{c}{L} \le 1$, repeated application of \eqref{eq23} gives the bound in \eqref{eq17} and the claimed limit.
\end{proof}
\noindent For stepsizes $\alpha_r = \alpha_1 2^{-r}$, the bound is
$\mathbb{E}[F(w_{k_r+1}) - F_*] \le 2F_{\alpha_r}$ and $ F_{\alpha_r} := \frac{\alpha_r L \tilde{M}}{2 c \mu_q}$ requiring $k_{r+1} - k_r \approx \frac{\log 3}{\alpha_r c \mu_q} = \mathcal{O}(2^r),$
with the Robbins–Monro condition $\sum_{k=1}^\infty \alpha_k = \infty$ and $\sum_{k=1}^\infty \alpha_k^2 < \infty.$

\begin{theorem}[Strongly Convex Objective, Diminishing Stepsizes with Quantization]
\label{theorem2}
Under Assumptions~\ref{assum1}, \ref{assum2}, and~\ref{assum3} (with $F_{\mathrm{inf}} = F_*$), suppose that SGD with the low-precision gradient $\tilde{g}(w_k,\xi_k)$ is run with a stepsize sequence $\alpha_k = \frac{\beta}{\gamma + k}, \beta > \frac{1}{c \mu_q}, \gamma > 0,$ such that $\alpha_1 \le \frac{\mu_q}{L \tilde{M}_G}$, where $\mu_q := q_{\min} \mu$.  
Then, for all $k \in \mathbb{N}$, the expected optimality gap satisfies
\begin{equation}
    \mathbb{E}[F(w_k) - F_*] \le \frac{\nu_q}{\gamma + k},
    \label{eq25}
\end{equation}
where $\nu_q := \max \left\{ \frac{\beta^2 L \tilde{M}}{2 (\beta c \mu_q - 1)}, \; (\gamma + 1)(F(w_1) - F_*) \right\}.$
Since $\mu_q$ appears in the denominator of $\frac{\beta^2 L \tilde{M}}{2 (\beta c \mu_q - 1)}$, a smaller $\mu_q$ reduces the denominator and thus increases the bound.
\end{theorem}

\begin{proof}
From the choice of $\alpha_k$, we have $\alpha_k L \tilde{M}_G \le \alpha_1 L \tilde{M}_G \le \mu_q$ for all $k \in \mathbb{N}$.  
Applying Lemma~\ref{lemma2} and the strong convexity property, for all $k \in \mathbb{N}$:
\begin{align}
    \mathbb{E}_{\xi_k}[F(w_{k+1})] - F(w_k)
    &\le -\left( \mu_q - \frac{1}{2} \alpha_k L \tilde{M}_G \right) \alpha_k \| \nabla F(w_k) \|_2^2
        + \frac{1}{2} \alpha_k^2 L \tilde{M} \label{eq27} \\
    &\le -\frac{1}{2} \alpha_k \mu_q \| \nabla F(w_k) \|_2^2
        + \frac{1}{2} \alpha_k^2 L \tilde{M} \label{eq28} \\
    &\le -\alpha_k c \mu_q \left( F(w_k) - F_* \right)
        + \frac{1}{2} \alpha_k^2 L \tilde{M}.
        \label{eq29}
\end{align}
Subtracting $F_*$ and taking expectations yields
\begin{equation}
    \mathbb{E}[F(w_{k+1}) - F_*]
    \le \underbrace{(1 - \alpha_k c \mu_q)}_{\text{stepsize shrinkage}} \, \mathbb{E}[F(w_k) - F_*] + \frac{1}{2} \alpha_k^2 L \tilde{M}.
    \label{eq30}
\end{equation}
We prove \eqref{eq25} by induction.  
For $k=1$, the definition of $\nu_q$ guarantees \eqref{eq25}.  
Assume \eqref{eq25} holds for some $k \ge 1$.  
Substitute into \eqref{eq30} with $\hat{k} := \gamma + k$:
\begin{align}
    \mathbb{E}[F(w_{k+1}) - F_*]
    &\le \left( 1 - \frac{\beta c \mu_q}{\hat{k}} \right) \frac{\nu_q}{\hat{k}} 
        + \frac{\beta^2 L \tilde{M}}{2 \hat{k}^2} \label{eq31} \\
    &= \frac{\hat{k} - \beta c \mu_q}{\hat{k}^2} \nu_q
        + \frac{\beta^2 L \tilde{M}}{2 \hat{k}^2} \label{eq32} \\
    &= \frac{\hat{k} - 1}{\hat{k}^2} \nu_q
        - \underbrace{\frac{\beta c \mu_q - 1}{\hat{k}^2} \nu_q + \frac{\beta^2 L \tilde{M}}{2 \hat{k}^2}}_{\le 0 \text{ by def. of }\nu_q}
        \label{eq33} \\
    &\le \frac{\nu_q}{\hat{k} + 1},
\end{align}
where the last inequality uses $\hat{k}^2 \ge (\hat{k} + 1)(\hat{k} - 1)$.  
Thus, \eqref{eq25} holds for $k+1$, completing the induction.
\end{proof}

Based on Theorem~\ref{theorem2}, when $q_{\min} < 1$, the reduced effective coefficient $\mu_q = q_{\min} \mu$ makes the $O(1/k)$ convergence rate slower and increases the constant factor in the bound.

\end{document}